\documentclass[10pt, conference, compsocconf]{IEEEtran}

\usepackage[pdftex]{graphicx}

\usepackage{graphicx}
\usepackage{amssymb}
\usepackage{amsmath}
\usepackage{amsfonts}
\usepackage{makeidx}
\usepackage{multirow}
\usepackage{verbatim}
\usepackage[usenames]{color}
\usepackage{algorithm}
\usepackage{algorithmic}

\begin{document}

 \begin{titlepage}
 \begin{center}
 {\Large \sc PREPRINT VERSION\\}
  \vspace{5mm}
{\huge Simplified firefly algorithm for 2D image key-points search\\}
 \vspace{10mm}
 {\Large C. Napoli, G. Pappalardo, E. Tramontana, Z. Marsza\l ek, D. Po\l ap and M. Wo\`zniak\\}
 \vspace{5mm}
{\Large \sc PUBLISHED ON: \bf 2014 IEEE Symposium on Computational Intelligence for Human-like Intelligence}
 \end{center}
 \vspace{5mm}
 {\Large \sc BIBITEX: \\}
 
@inproceedings\{Napoli2014Simplified,\\
year=\{2014\},\\
isbn=\{978-1-4799-4507-8\},\\
booktitle=\{2014 IEEE Symposium on Computational Intelligence for Human-like Intelligence\},\\
title=\{Simplified firefly algorithm for 2D image key-points search\}, \\
publisher=\{IEEE\},\\
author=\{Napoli, Christian and Pappalardo, Giuseppe and Tramontana, Emiliano and Marsza\l ek, Zbigniew and Po\l ap, David and Wo\`zniak, Marcin\},\\
month=\{December 9-12\},\\
address=\{Orlando, Florida, USA\},\\
pages=\{118-125\}\\
\}
 \vspace{5mm}
 \begin{center}
Published version copyright \copyright~2014 IEEE \\
\vspace{5mm}
UPLOADED UNDER SELF-ARCHIVING POLICIES\\
NO COPYRIGHT INFRINGEMENT INTENDED \\
 \end{center}
\end{titlepage}

\title{Simplified firefly algorithm for 2D image key-points search}
\author{\IEEEauthorblockN{Christian Napoli\IEEEauthorrefmark{2}, Giuseppe Pappalardo\IEEEauthorrefmark{2}, Emiliano Tramontana\IEEEauthorrefmark{2},  Zbigniew Marsza{\l}ek\IEEEauthorrefmark{1}, Dawid Po{\l}ap\IEEEauthorrefmark{1}\\
and Marcin Wo{\'z}niak\IEEEauthorrefmark{1} 
}

\IEEEauthorblockA{\IEEEauthorrefmark{2}Department of Mathematics and Informatics, University of Catania,\\
Viale A. Doria 6, 95125 Catania, Italy\\
Email: Napoli@dmi.unict.it, Pappalardo@dmi.unict.it, Tramontana@dmi.unict.it}

\IEEEauthorblockA{\IEEEauthorrefmark{1}Institute of Mathematics, Silesian University of Technology,\\
Kaszubska 23, 44-100 Gliwice, Poland\\ Email: Zbigniew.Marszalek@polsl.pl, Dawid.Polap@gmail.com,  Marcin.Wozniak@polsl.pl}
}
\maketitle


\begin{abstract}
In order to  identify an object, human eyes firstly search the field of view for points or areas which have particular properties. These properties are used to recognise an image or an object. Then this process could be taken as a model to develop computer algorithms for images identification.
This paper proposes the idea of applying the simplified firefly algorithm to search for key-areas in 2D images. For a set of input test images the proposed version of firefly algorithm has been examined. Research results are presented and discussed to show the efficiency of this evolutionary computation method.
\end{abstract}


%
\IEEEpeerreviewmaketitle

\section{Introduction}\label{sec:intro}
In modern computer science, evolutionary computation (EC) is one of most important fields, widely applied in various tasks. There are many applications of EC in sciences and industry. The power of computational intelligence (CI) with dedicated mechanisms is used to simulate even sophisticated phenomenon. EC is efficient for searching optimal solutions, easy to implement and precise. Let us give same examples.

EC applied to create learning sets for artificial intelligence (AI) control systems is discussed in \cite{wozniak2012,capizzi2012innovative,napoli2010hybrid}. Some aspects of positioning computing network models by the use of EC are presented in \cite{wozniak2013_1,wozniak2013_2,napoli2010exploiting} and \cite{wozniak2014_1,napoli2013hybrid} or \cite{wozniak2014_2}. In \cite{wozniak2008_3}, \cite{wozniak2008_2} and \cite{wozniak2009, BannoMPT10b} applications of EC methods in dynamic systems positioning and simulation is presented. Optimization of industry processes, i.e. iron cast simulation is shown in \cite{hetmaniok2013}. EC methods are also supposed to be more adaptive and efficient in comparison to classic optimization methods, see \cite{cabello2009,capizzi2011hybrid} and \cite{hu2013,bonanno2012optimal}. Summing up, EC methods are applied where CI may help to improve data processing. All this gave some inspiration to implement dedicated EC in 2D image processing. 

First attempts to apply CI methods in 2D image processing are discussed in \cite{Wozniak2014_3}, \cite{Wozniak2014_5} and \cite{Wozniak2014_6}. In \cite{Wozniak2014_3,tangling13} some aspects of handwriting preprocessing for artificial intelligence (AI) classification systems are discussed. \cite{Wozniak2014_5} and \cite{Wozniak2014_6} show ideas of a novel approach to identify simple objects in 2D images by the use of EC methods. These results can help in move methods refactoring of large systems (please see \cite{NapoliCISIS,GiuntaPT12,PappalardoT13}) and surface plasmon polaritons in thin metals (please see \cite{NapoliICAISC}), where we need to identify input objects, as well as an advanced modelling technique for a wide variety of systems \cite{bonanno2012some,capizzi2011recurrent,capizzi2010new}.

In this paper we discuss potential efficiency of dedicated firefly algorithm (FA) to search for key areas in 2D input objects. The purpose of the developed algorithm is to find a advanced solution to obtain precise whole-areas recognition by means of a few iterations. Moreover the proposed FA is designed for better efficiency with respect to other similar recognition algorithms, while still being easy to implement thanks to its simplicity.

The rest of this paper is structured as follows.  Section~\ref{sec:objectmodeling} 
 gives the background on the Key-Points search and some classical approaches, then introduces the proposed FA solution. Section~\ref{sec:auxiliaryresults} describes the implementation of the proposed FA and presents the experimental results.  Finally, Section~\ref{sec:Final}
draws our conclusions.

\section{Key-Points search}\label{sec:objectmodeling}
\begin{figure}[tp]
\centerline{\includegraphics[width=3.5in]{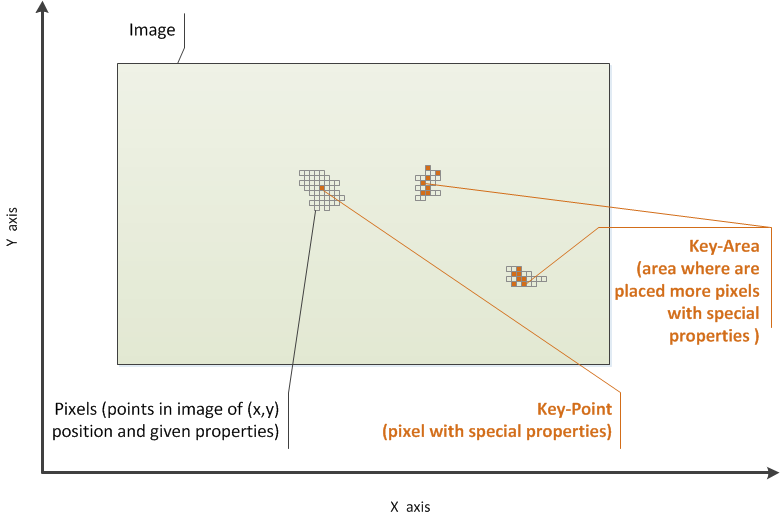}} \caption{Schematic key-points position in 2D input image} \label{fig:key_points}
\end{figure}

In computer image, each classified object consists of points, which have special position and properties. Among them one can name saturation, sharpness, brightness and more. All these features compose an image of the input object that is visible to our eyes. An exemplary model is shown in Fig.  \ref{fig:key_points}.
By analysing such features it is possible to identify objects. However recognising objects depends on the selection of meaningful key parts of the image.  Therefore one may say, that the position of each pixel (each one has measurable coordinates $\mathbf{x}=(x,y)=(x_{i,1},x_{i,2})$) and it's properties (like brightness) are important aspects to make right decision. We decide to recognise some parts or important areas, which can be defined for a computer system. Thus, \textit{Key-Point} is a pixel in 2D input image, which has peculiar properties making it important for object recognition. \textit{Key-Area} is containing many key-points that all together compose an object to recognize. In other words, to find the object our eyes search for areas in the picture that contain many points of the same kind. They compose an image of the object, which is being recognized in our brain. This process has some features that may be implemented in a computer algorithm. 
CI brings many interesting methods that can help. It is possible to apply EC method, in particular simplified FA, to perform the process of searching key-areas which compose input objects we are looking for. In our experiments we take sample images from open test images database\footnote{\textit{www.imageprocessingplace.com}} to examine. The performed experiments show potential efficiency of dedicated FA in the search for key-areas in various 2D pictures. This makes the presented solution not only efficient but also much easier to implement in comparison to classic methods.
\ref{fig:key_points}. 

\subsection{SURF - Classic Attempt}\label{sec:SURFalgorithm}
The classic methods used for key-areas recognition is SURF (Speeded-Up Robust Features) algorithm.  This method describe the input image by selecting characteristic key-points. Let us give only a short description here, for more details please see \cite{abeles2013}, \cite{bay2008}, \cite{decker2011_1}, \cite{gossow2011} or \cite{mehrotra2009}. In this work we used SURF combining selection of key-points with calculating 64-element vector (descriptor). In the SURF it is applied integrated image and filter approximation of block Hessian determinant. To detect interesting points we used particular Hessian-matrix approximation (for details please see \cite{abeles2013} and \cite{bay2008}).

The implementd SURF algorithm define Hessian matrix $H(\mathbf{x},\sigma)$ for point $\mathbf{x}=(x,y)$ in 2D image $I$  at scale $\sigma$ using formula
\begin{equation}\label{hessianvalue}
	H(\mathbf{x},\sigma)=
	\begin{bmatrix}
		L_{xx}(\mathbf{x},\sigma)&L_{xy}(\mathbf{x},\sigma)\\
		L_{xy}(\mathbf{x},\sigma)&L_{yy}(\mathbf{x},\sigma)
	\end{bmatrix},
\end{equation}
where the notations are: $L_{xx}(\mathbf{x},\sigma)$ -- convolution of Gaussian second order derivative $\frac{\partial^2}{\partial x^2}$. We define approximation $D_{xx}$, $D_{yy}$ and $D_{xy}$ using formula
\begin{equation}
		det(H_{ap})=D_{xx}D_{yy}-\left(\frac{|L_{xy}(\sigma)|_F|D_{xx}(\sigma)|_F}{|L_{xx}(\sigma)|_F|D_{xy}(\sigma)|_F}D_{xy}\right)^2.
\end{equation}
Then, the input image is blurred to get DoG (Difference of Gaussian) images, which help to find edges. To localize interesting points it is used non--maximum suppression in $3\times 3\times 3$ neighborhood (for more results please see also \cite{Wozniak2014_5} and \cite{Wozniak2014_6}). Maximum determinant of Hessian matrix is interpolated at scale $\sigma$, which helps to differ between first level and each octave. The applied SURF for RGB (Red-Green-Blue) color values model is presented in algorithm \ref{alg:SURF}.
\begin{algorithm}[!ht]
\caption{The SURF method applied to search for key-points in 2D images}
\label{alg:SURF}
\begin{algorithmic}[1]
\STATE Calculate number of $pixels$ and $rows$ in 2D image,
\WHILE{$j < rows$}
\WHILE{ $i < pixels$}
\STATE $rowsum_j=(R_{ij})^2+(G_{ij})^2+(B_{ij})^2$,
\ENDWHILE
\ENDWHILE
\STATE Integral image is sum of $rowsum$,
\STATE Calculate approximated Hessian value using (\ref{hessianvalue}), 
\STATE Build response layers for the image,
\WHILE{ $i < pixels$}
\STATE Calculate descriptor vector for each point,
\STATE Determine orientation,
\ENDWHILE
\STATE Construct descriptor vector for each image point.
\end{algorithmic}
\end{algorithm}
\subsection{SIFT - Classic Attempt}\label{sec:SIFTalgorithm}
SIFT (Scale-Invariant Feature Transform) transforms image into scale-invariant coordinates relative to local aspects. This generates features that may densely cover image for full range of scales and locations, please see \cite{Pope1998} and \cite{Se2002}. SIFT idea is based on \cite{Zhang1995}, where possibility of matching Harris corners over large image by using correlation window around each corner was discussed (see also \cite{Azad2009} and \cite{Sun2013} for details). This idea was developed in \cite{Schmid1997} to general image recognition, where Harris corners were applied to select key-points by rotationally invariant descriptor of local image regions. However Harris corner detector can be sensitive to changes in image scale, what makes it inefficient in processing images of different sizes. Therefore \cite{Nelson1998}, \cite{Pope1998} and \cite{Shokoufandeh1999} extended local feature approach to achieve scale invariance. Then some special features like multidimensional histograms summarizing distribution of measurements useful for recognition of textured objects with deformable shapes were discussed in \cite{Schiele2000}. Final version of this scale descriptor, less sensitive to local image distortions, was given in \cite{Lowe2004}. 

In the implemented SIFT, features are first extracted from set of images and stored in memory. Key-point are matched by individually compared examined feature to these previously stored using Euclidean distance of feature vectors. Correct key-points are filtered from set of matches by identifying subsets of interest points that agree on object, it's location, scale and orientation. To determine these clusters we perform hash table implementation of generalized Hough transform. Each cluster of features is then subject to further verification. SIFT uses special detector to find scale-space extrema where continuous function is Gaussian. According to \cite{Lowe2004} we have used $L(X,\gamma_D)$ as scale space of image defined in
\begin{equation}
 L(X,\gamma_D)=G(X,\gamma_D)*I(X),
\end{equation}
where $L(X,\gamma_D)$ is produced from convolution of variable-scale
Gaussian $G(X,\gamma_D)$ with input image point. To scale selection is used approximated DoG filter. Then key-point is localized by taking Taylor series expansion of scale-space function
\begin{equation}
D(\vec{X})=D+D^T_{x}\vec{X}+0.5\vec{X^T}D^T_{xx}\vec{X},
\end{equation}
where $D(\vec{X})$ and it's derivatives are evaluated at image points and $T$ is offset from these points (for details see \cite{Brown2002}). Finally after filtering, descriptor operations are performed. Descriptor is local statistic of orientations of the gradient of the Gaussian scale space. 
\begin{algorithm}[!ht]
\caption{Simplified SIFT applied to search for key-points in 2D images}
\label{alg:SIFT}
\begin{algorithmic}[1]
\STATE Calculate maximum number of $Level$,
\STATE Build DoG pyramid,
\STATE Find $maximum$ and $minimum$ of $Level$,
\STATE Compare vector of pixel with it's rescaled neighbors,
\FOR{Each $Level$ of DoG pyramid}
    \STATE Match to sub pixel $maximum$ location,
    \STATE Eliminate edge points,
\ENDFOR
\STATE Construct keys using interpolated value.
\end{algorithmic}
\end{algorithm}  
\subsection{Simplified Firefly Algorithm - A Novel Approach}\label{sec:FAalgorithm}
Very efficient methods that can be applied in the process of key-areas search are EC algorithms. One of them is simplified firefly algorithm (SFA). It is mapping behavior of flying and blinking insects while searching for a partner. This process, in our simplified version, is applied in 2D images key-areas search. 

Since classic version of FA, first presented in \cite{yang2010_2} and \cite{yang2009}, it was applied in many fields. In \cite{coelho2011} chaotic FA was applied in reliability-redundancy optimization. In \cite{horng2011} it is applied to minimum cross entropy threshold selection. In \cite{jati2011} and \cite{yousif2011} this method is applied to solve traveling salesman problem and jobs scheduling. It is also efficient in continuous optimization (see \cite{yang2013}) or multi-modal optimization (see \cite{yang2010_1}). While in \cite{horng2012_1} is presented efficient application of FA in vector quantization. FA is also efficient in neural network training and power systems positioning (see \cite{nandy2012} and \cite{rampriya2010}, respectively). 
Finally some important aspects of applying FA in the process of image compression are described in \cite{horng2011} and \cite{Wozniak2014_5}. Where in \cite{Wozniak2014_5} was discussed idea for potential application of FA classic version in 2D image processing.
Therefore here we present SFA method devoted to 2D image processing. Let us now present mathematical model of the SFA algorithm. 

The SFA is mapping behavior of fireflies in natural conditions. Individuals are described by several biological traits: specific way of flashing, specific way of moving and specific perception of the others. These are mathematically modeled in implementation of SFA as:
\begin{itemize}
\item $\gamma$--light absorption coefficient in given circumstances,
\item $\mu$--firefly random motion factor,
\item $\beta_{pop}$--firefly attractiveness factor,
\end{itemize}
which implement behavior of different species of fireflies and natural conditions of the environment. Just as in nature, a firefly goes to the most attractive other one by measuring the intensity of flickers over the distance between them characterized by a suitable metric. In SFA an average distance $r_{ij}$ between any two fireflies $i$ and $j$ maps the inverse square law. Attractiveness of individuals decreases with increasing distance $r_{ij}$ between them. We also map air absorption of light, which makes fireflies visible to certain distance. Using these we built CI to map the behavior of fireflies. In description of SFA we assume:
\begin{itemize}
\item All fireflies are unisex, therefore one individual can be attracted to any other firefly regardless of gender, 
\item Attractiveness is proportional to brightness. Thus, for every two fireflies less clear flashing one will move toward brighter one,
\item Attractiveness decreases with increasing distance between individuals,
\item If there is no clearer and more visible firefly within the range, then each one will move randomly. 
\end{itemize}

Distance between any two fireflies $i$ and $j$ situated at points $\mathbf{x}_i$ and $\mathbf{x}_j$ in 2D image we define using Cartesian metric
\begin{equation}\label{FAdistance}
r_{ij}^{t}=\lVert \mathbf{x}_i^{t} - \mathbf{x}_j^{t} \rVert = \sqrt{\sum^{2}_{k=1}(x_{i,k}^{t}-x_{j,k}^{t})^2},
\end{equation}
where notations in $t$ iteration are: $\mathbf{x}_i^{t}$, $\mathbf{x}_j^{t}$--points in $R\times R$ space (here pixels coefficients on axis X and Y - see Fig. \ref{fig:key_points}), $x_{i,k}^{t}$, $x_{k,j}^{t}$--k-th components of the spatial coordinates $\mathbf{x}_i^{t}$ and $\mathbf{x}_j^{t}$ that describe each firefly (2D image point) in the space.

Attractiveness of firefly $i$ to firefly $j$ decreases with increasing distance. Attractiveness is proportional to intensity of light seen by surrounding individuals and defined as
\begin{equation}\label{FAatractiveness}
\beta_{ij}(r_{ij}^{t})=\beta_{pop}\cdot e^{-\gamma \cdot (r_{ij}^{t})^2},
\end{equation}
where notations in $t$ iteration are: $\beta_{ij}(r_{ij}^{t})$--attractiveness of firefly $i$ to firefly $j$, $r_{ij}^{t}$--distance between firefly $i$ and firefly $j$, $\gamma$--light absorption factor mapping natural conditions, $\beta_{pop}$--firefly attractiveness factor.

Firefly $i$ motions toward more attractive and clearer flashing individual $j$ using information about other individuals in the population denotes simplified formula
\begin{equation}\label{FAmove}
\mathbf{x}_{i}^{t+1} = \lfloor \mathbf{x}_{i}^{t}+\beta_{ij}(r_{ij}^{t})\cdot(\mathbf{x}_j^{t}-\mathbf{x}_i^{t})+\mu e_i \rfloor,
\end{equation}
where notations in $t$ iteration are: $\mathbf{x}_i^{t}$, $\mathbf{x}_j^{t}$--points in $R \times R$ space (pixels coefficients on axis X and Y), $r_{ij}^{t}$--distance between fireflies $i$ and $j$ modeled in (\ref{FAdistance}), $\beta_{ij}(r_{ij}^{t})$--attractiveness of firefly $i$ to firefly $j$ modeled in (\ref{FAatractiveness}), $\mu$--coefficient mapping natural random motion of fireflies, $e_i$--randomized vector changing position of firefly on each axis. Formula (\ref{FAmove}) is dedicated and simplified version for 2D images. We introduce it for 2D image key-area search. Using only integer values for X and Y coefficients we can precisely move from pixel to pixel. SFA implementation is presented in algorithm \ref{alg:FAalgorithm}.
\begin{algorithm}[!ht]
\caption{Simplified FA to search for 2D image key-points}
\label{alg:FAalgorithm}
\begin{algorithmic}[1]
\STATE Define all coefficients: $\gamma$--light absorption factor, $\beta_{pop}$--attractiveness factor, $\mu$--natural random motion factor, number of $fireflies$ and $generation$--number of iterations in the algorithm,
\STATE	Define fitness function for the algorithm -- properties of pixels to search for according to (\ref{Brightness}),
\STATE	Create at random initial population of $fireflies$ in the picture,
\STATE t:=0,			
	\WHILE{$t \leq generation$}
		\STATE Calculate distance between individuals in population $P$ using (\ref{FAdistance}),
		\STATE Calculate attractiveness for individuals in population $P$ using (\ref{FAatractiveness}),
		\STATE Evaluate individuals in population $P$ using (\ref{Brightness}),
		\STATE Create population $O$: move individuals towards closest and most attractive individual using (\ref{FAmove}),
		\STATE Evaluate individuals in population $O$ using (\ref{Brightness}),
		\STATE Replace $best\_ratio$ individuals from population $P$ with $best\_ratio$ individuals from population $O$, the rest take at random,
		\STATE Rest of $fireflies$ take at random,
		\STATE Next $generation$ $t:=t+1$,
	\ENDWHILE
\STATE Best $fireflies$ from the last $generation$ in population $P$ are potential key-points in 2D image.
\end{algorithmic}
\end{algorithm}
\section{Research Results}\label{sec:auxiliaryresults}
SFA was applied to search for 2D image key-areas. Each firefly is representing a single pixel (point in 2D input image according to coordinates in Fig. \ref{fig:key_points}). Therefore in this paper we simultaneously change names: pixel, 2D image point and firefly. In each iteration we move entire population to search between all image points. Fireflies move from pixel to pixel and search for specific areas according to a given criterion. In our experiments we have used simplified fitness function, which reflects brightness and sharpness of the input image points
\begin{equation}\label{Brightness} 
\Phi(\mathbf{x}_i)=\Phi((x_{i,1},x_{i,2}))=\left\lbrace 
\begin{array}{ll}
0.1 \dots 1 & \text{ saturation}\\
0 & \text{other}\\
\end{array} \right. ,
\end{equation}
where notation $\Phi(\mathbf{x}_i)$ denotes quality of evaluated pixel. This measure reflects a value in the range from $0.0$ to $1.0$, where color saturation change from black to white. Therefore using SFA with simple fitness function we are to build SFA classifier based on canny or sobel filter (for details on 2D image filtering please see \cite{Pratt2001}). Using filtering we extract borders of input objects, which will be marked in white on dark background. Therefore these filtered 2D images will be perfect input objects for final SFA classifier. However, here we would like to discuss only the first part of the project. It is SFA efficiency in classifying key-areas in 2D input images.

When fireflies fly for each iterations, they pick pixels with best fitness within the range of their flight. They take position and wait for classification. From all the individuals we take $best\_ratio$ of them with highest or lowest fitness function value (depending on the experiment). These points (fireflies) are taken to next generation. The rest of population is randomly placed among all input image points, what helps to search entire input object for the points of interest. 

Let us present results of search for quality of pixel defined in (\ref{Brightness}), using simplified SURF from section \ref{sec:SURFalgorithm}, SIFT from section \ref{sec:SIFTalgorithm} and SFA from section \ref{sec:FAalgorithm}. Simulations were performed for $400$ fireflies in $20$ generations with set: $\beta_{pop}=0.3$, $\gamma=0.3$, $\mu=0.25$, $best\_ratio=35\%$. We have examined SFA on standard test images downloaded from open test image databases (see section \ref{sec:intro}). Experiments were performed on various types of 2D input pictures: sharp, blurred, landscapes and human postures or faces. Each of resulted key-points (pixels) is marked in red. We have provided some close-ups of classified areas for better presentation. First we present dark areas where $\Phi(\mathbf{x}_i)=0$ or $\Phi(\mathbf{x}_i)=0.1$. In second attempt we present bright areas where $\Phi(\mathbf{x}_i)=0.9$ or $\Phi(\mathbf{x}_i)=1$. 
\subsection{Dark areas in 2D images}\label{sec:dark}
Let us first present research results for dark objects localization. Dark objects are present in many different images. They can represent objects in landscape (trees, blocks, different constructions, etc.), natural phenomena (tornadoes, shadows, etc.), human figures or human appearance (face features, hair, eyes, etc.). 
\begin{figure}[!h]
\centerline{
\includegraphics[width=1.8in]{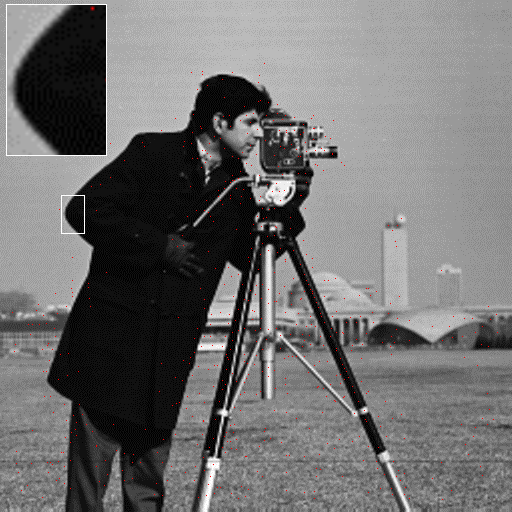}
\includegraphics[width=1.8in]{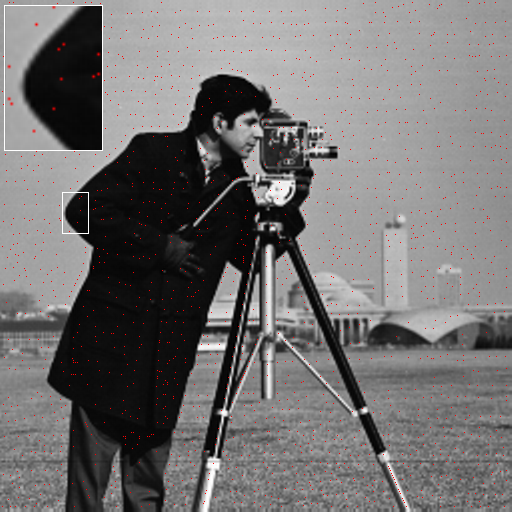}
} \text{   }  \newline
\centerline{
\includegraphics[width=1.8in]{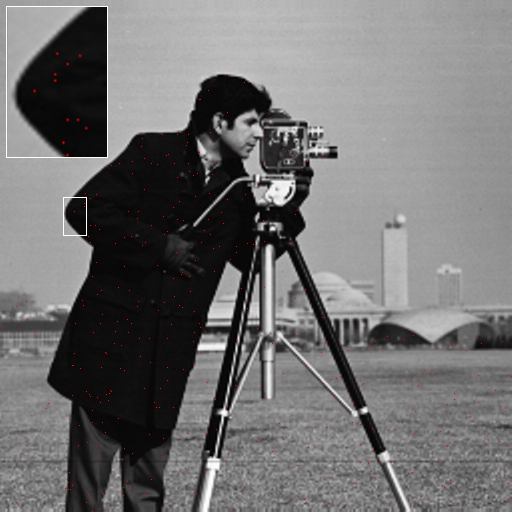}
}
\caption{Dark key-points search in sharp and human posture images: in first row on the left simplified SURF result and on the right simplified SIFT result, in the second row SFA result} \label{fig:humanposture_black_keypoints}
\end{figure}
\begin{figure}[!h]
\centerline{
\includegraphics[width=1.8in]{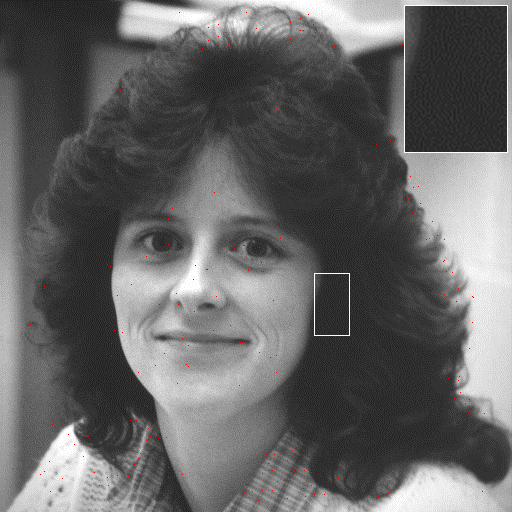}
\includegraphics[width=1.8in]{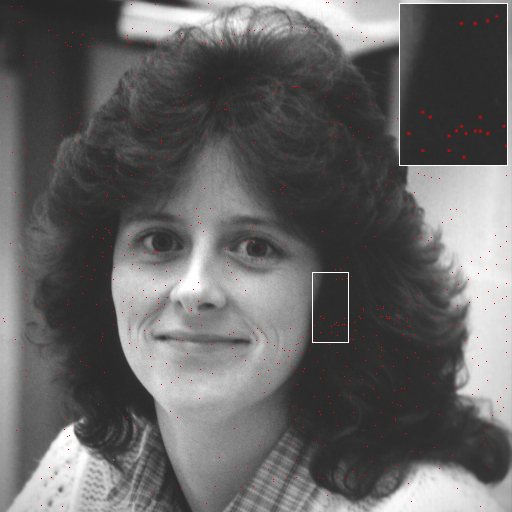}
}\text{   }  \newline
\centerline{
\includegraphics[width=1.8in]{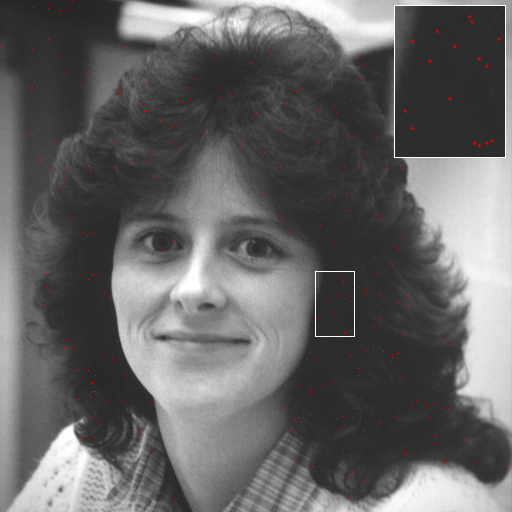}
}
\caption{Dark key-points search in sharp and human face images: in first row on the left simplified SURF result and on the right simplified SIFT result, in the second row SFA result} \label{fig:face_black_keypoints}
\end{figure}
\begin{figure}[!h]
\centerline{
\includegraphics[width=1.8in]{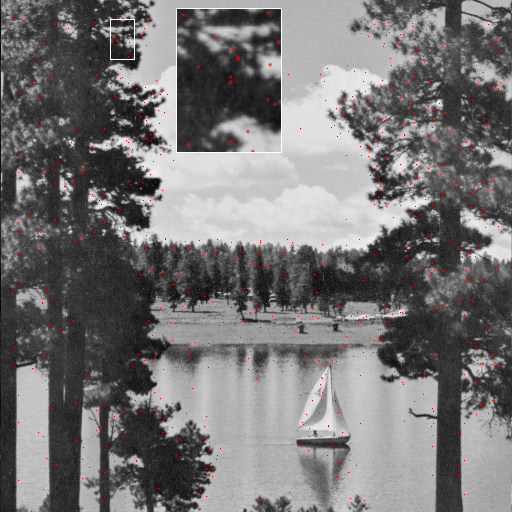}
\includegraphics[width=1.8in]{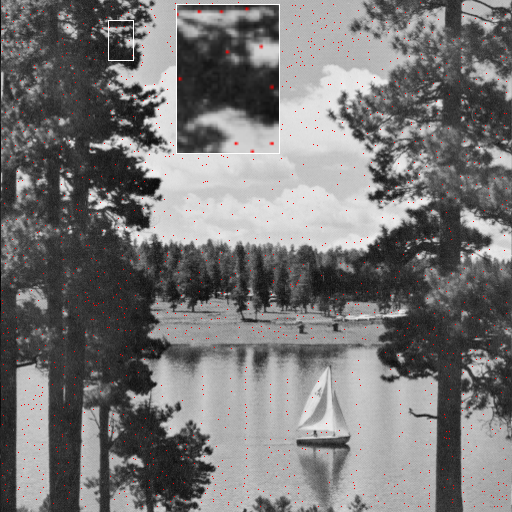}
}\text{   }  \newline
\centerline{
\includegraphics[width=1.8in]{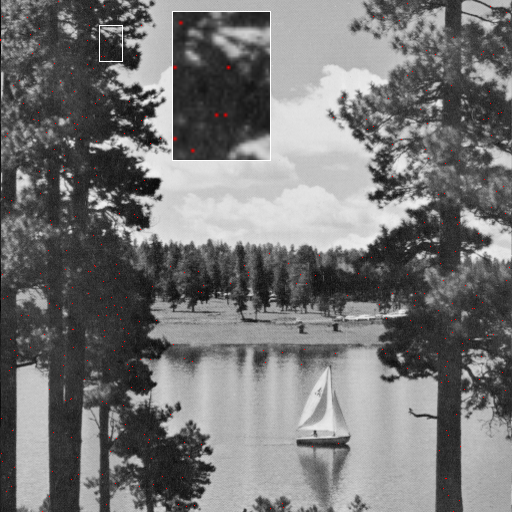}
}
\caption{Dark key-points search in landscape images: in first row on the left simplified SURF result and on the right simplified SIFT result, in the second row SFA result} \label{fig:landscape1_black_keypoints}
\end{figure}
\begin{figure}[!h]
\centerline{
\includegraphics[width=1.8in]{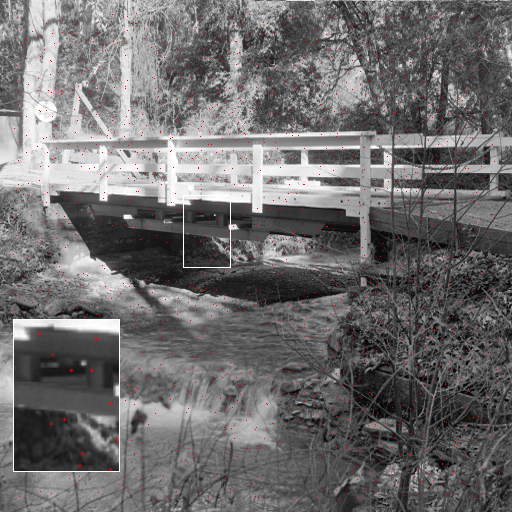}
\includegraphics[width=1.8in]{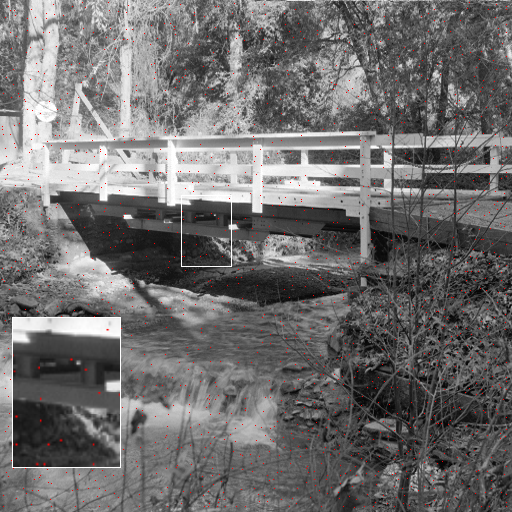}
}\text{   }  \newline
\centerline{
\includegraphics[width=1.8in]{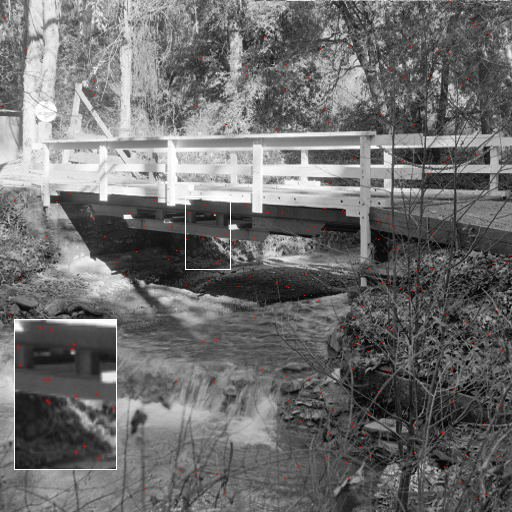}
} \caption{Dark key-points search in sharp landscape images: in first row on the left simplified SURF result and on the right simplified SIFT result, in the second row SFA result} \label{fig:landscape2_black_keypoints}
\end{figure}
\begin{figure}[!h]
\centerline{
\includegraphics[width=1.8in]{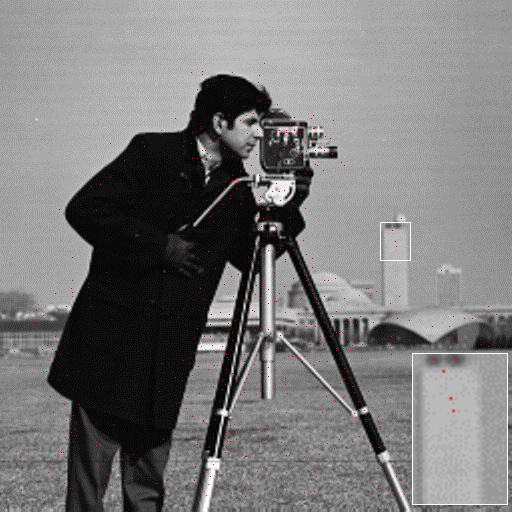}
\includegraphics[width=1.8in]{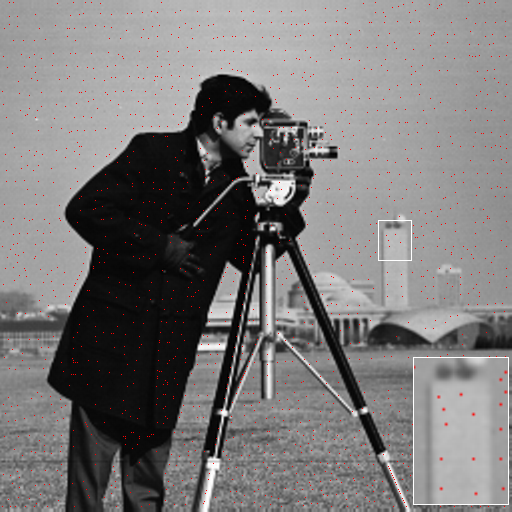}
}\text{   }  \newline
\centerline{
\includegraphics[width=1.8in]{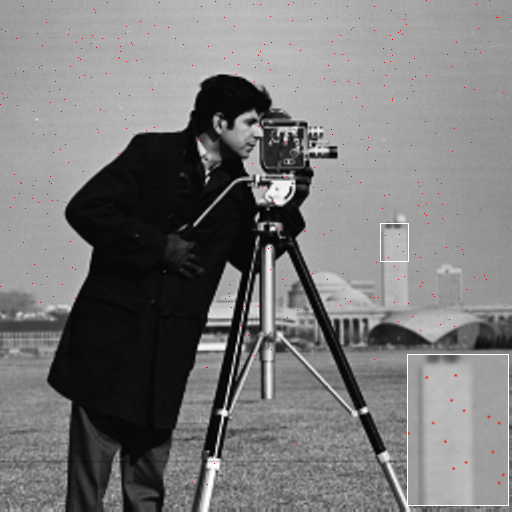}
} 
\caption{Bright key-points search in sharp and human posture images: in first row on the left simplified SURF result and on the right simplified SIFT result, in the second row SFA result} \label{fig:humanposture_bright_keypoints}
\end{figure}
\begin{figure}[!h]
\centerline{
\includegraphics[width=1.8in]{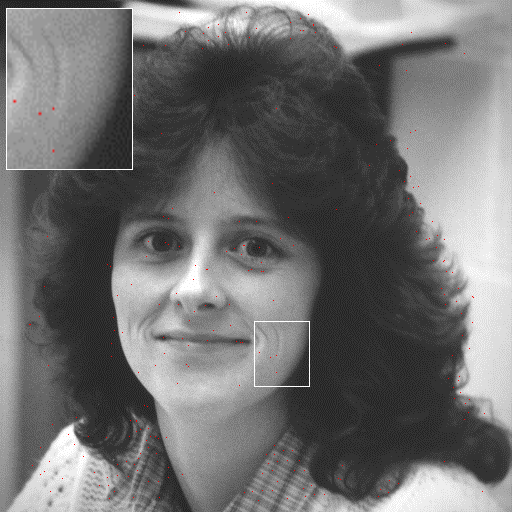}
\includegraphics[width=1.8in]{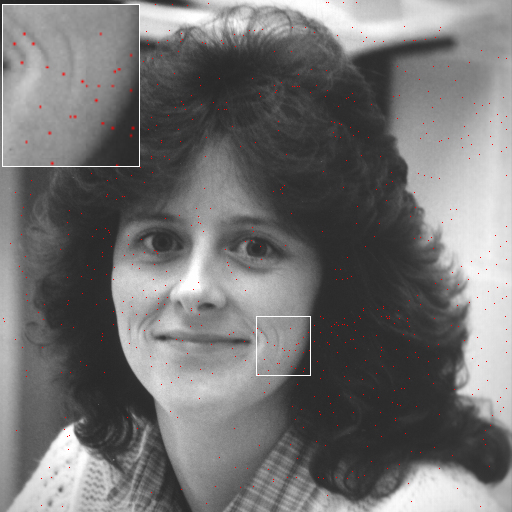}
}\text{   }  \newline
\centerline{
\includegraphics[width=1.8in]{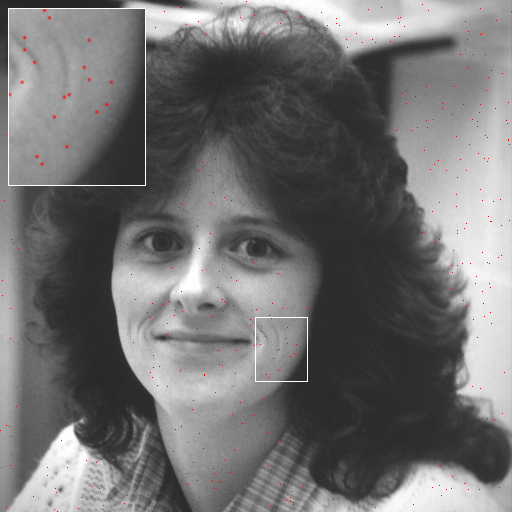}
}
\caption{Bright key-points search in sharp and human face images: in first row on the left simplified SURF result and on the right simplified SIFT result, in the second row SFA result} \label{fig:face_bright_keypoints}
\end{figure}
\begin{figure}[!h]
\centerline{
\includegraphics[width=1.8in]{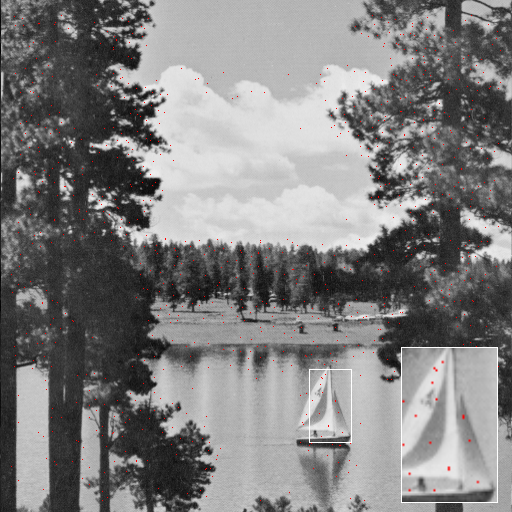}
\includegraphics[width=1.8in]{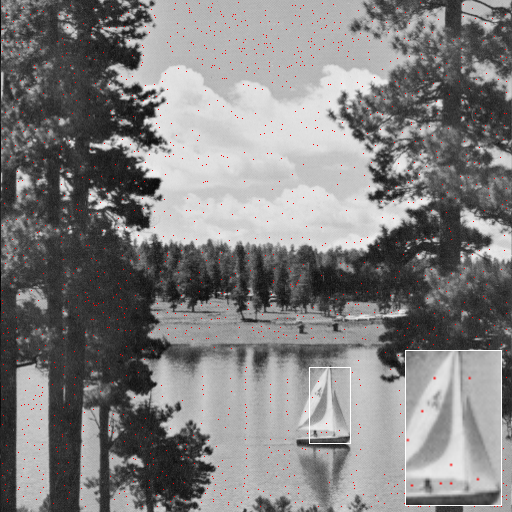}
}\text{   }  \newline
\centerline{
\includegraphics[width=1.8in]{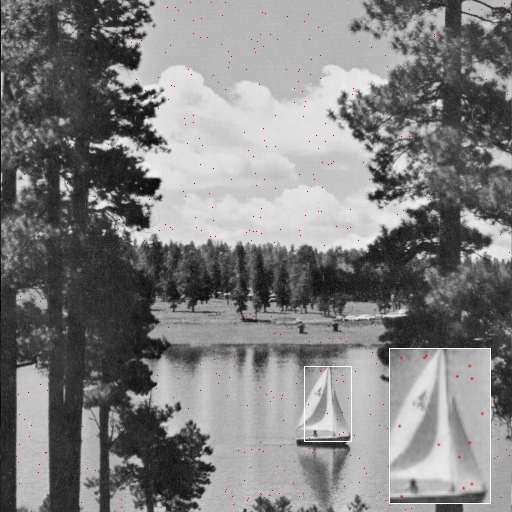}
}
\caption{Bright key-points search in landscape images: on the left SURF result and on the right simplified FA result} \label{fig:landscape1_bright_keypoints}
\end{figure}
\begin{figure}[!h]
\centerline{
\includegraphics[width=1.8in]{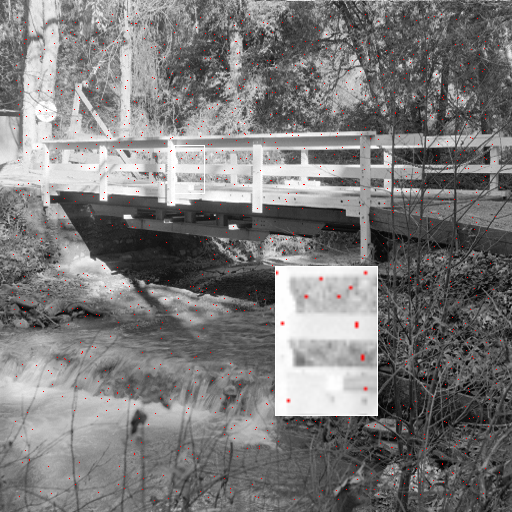}
\includegraphics[width=1.8in]{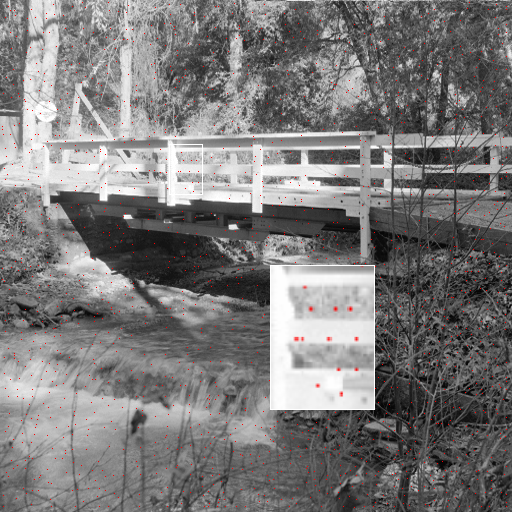}
}\text{   }  \newline
\centerline{
\includegraphics[width=1.8in]{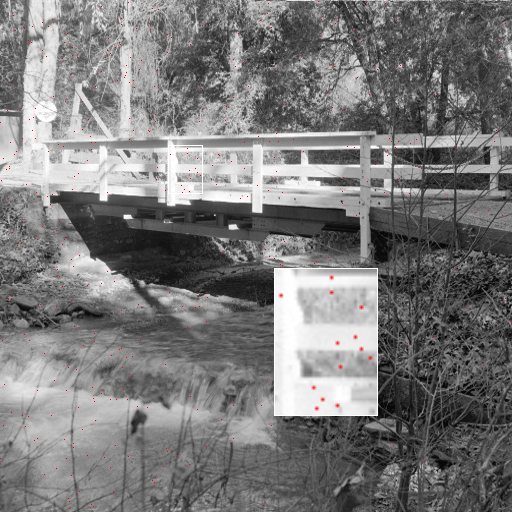}
} \caption{Bright key-points search in sharp landscape images: in first row on the left simplified SURF result and on the right simplified SIFT result, in the second row SFA result} \label{fig:landscape2_bright_keypoints}
\end{figure}

In Fig.\ref{fig:humanposture_black_keypoints} -- Fig.\ref{fig:landscape2_black_keypoints} are presented research results for dark key-areas in input images. We can see that SFA can easily find dark objects of different shapes like human posture, human appearance features (hair, eyes, etc.). It is also efficient in finding some aspects in landscapes. In Fig.\ref{fig:landscape1_black_keypoints} and Fig.\ref{fig:landscape2_black_keypoints} are presented research results of searching for shades under trees, buildings, bridges or some other natural phenomena. All these areas were found by SFA correctly.
\subsection{Bright areas in 2D images}\label{sec:bright}
Let us now present research results for bright objects localization. Bright objects are present in various images. They can represent landscape (bright or lightened constructions, boats, sails, etc.), natural phenomena (clouds, sun, stars, etc.), human figures or human appearance (gray hair, eyes, make-up, bright clothing, etc.). 

In Fig.\ref{fig:humanposture_bright_keypoints} -- Fig.\ref{fig:landscape2_bright_keypoints} are presented research results for bright key-areas in 2D input images. We can see that SFA can find bright objects of various shapes like human faces or bright clothes (see Fig. \ref{fig:face_bright_keypoints}). It is also efficient in locating bright or lightened constructions like bridges or buildings present in Fig.\ref{fig:humanposture_bright_keypoints} and Fig.\ref{fig:landscape2_bright_keypoints}. If compared to SURF, SFA results are better. This method may find more areas that correspond to given criterion in shorter time. 

This task is more complicated than searching for dark areas. Here the SFA must find bright points among many pixels of similar kind. As the photos were taken during day and all objects of bright properties are lightened in some way. Therefore calculations are slightly more complex. For example in Fig.\ref{fig:landscape1_bright_keypoints} we were looking for bright constructions like sails or boats located on water, which was also lightened by the Sun. Moreover there were also some clouds on the sky. All these features made processing more complicated. However this can be used as an advance in SFA final classifier based on canny and sobel filter. If applied to filtered 2D images, SFA operates on objects which are filtered and we see only white borders. Therefore recognition process will be easy to perform. 

\subsection{Conclusions}\label{sec:conclusions}
Application of SFA allows us to easily and reliably find key-areas in examined 2D input images. SFA is efficient when applied to search for areas like human postures, human face appearance (hair or eyes), building or nature elements (trees, dark constructions or nature phenomena like shades). Table \ref{tab:tablecomputetime} presents comparison and assessments of the examined methods. The grade was given in a scale from - - (what represents weak grade) to + + (what represents high grade). 
\begin{table}[!h]
\caption{Comparison and assessment}
\label{tab:tablecomputetime}
\centering
\begin{tabular}{|ccc|}
\hline
Verified feature & Examined method & Grade\\
\hline
\hline
 & SURF & -\\
Easy to implement & SIFT & -\\
 & SFA & +\\
\hline
\hline
 & SURF & + +\\
Fast recognition & SIFT & +/-\\
 & SFA & +/-\\
\hline
\hline
  & SURF & - -\\
Simple mathematic operations & SIFT & - -\\
  & SFA & + +\\
\hline
\hline
 & SURF & +\\
Precision in border recognition & SIFT & + +\\
 & SFA & +\\
\hline
\hline
 & SURF & +\\
Precision in whole area recognition & SIFT & +\\
 & SFA & + +\\
\hline
\hline
 & SURF & +\\
Precision using low number of iterations & SIFT & -\\
 & SFA & + +\\
\hline
\hline
 & SURF & +/-\\
Possible to increase efficiency & SIFT & +/-\\
 & SFA & + +\\
\hline
\end{tabular}
\end{table}
Comparing research results we conclude that calculations performed by SFA are simple. We just use formulas (\ref{FAdistance}) -- (\ref{FAmove}) to calculate position and perform move of each point in examined 2D input images. SURF and SIFT may present better precision in recognition of input object border line recognition, as it is one of their original purposes. SFA is covering whole recognized objects with points. However SFA, in final 2D classifier will be applied to search for certain objects in 2D input images, therefore we can think of it as a main advantage. Moreover SFA algorithm efficiency is increased if we are looking for key-points with high contrast in relation to surroundings, what will be used in final 2D recognition system based on canny and sobel filter for SFA input images.
\section{Final Remarks}
\label{sec:Final}
Research presented in this paper show that EC methods are an excellent tool to perform process of key-areas search in 2D images of any kind. It is possible to improve object recognition by application of sophisticated fitness function, which can precisely describe properties of objects and introduction of image filtering. In presented experiments we were looking for sample objects. Recognized points present various, miscellaneous but hypothetical cases. However, in the real applications, identification of objects in 2D images becomes nontrivial problem. Search objects can be not easy to recognize, as discussed for some examples in section \ref{sec:bright} due to e.g. wear (expenditure) of elements not enough contrast and environment conditions (lightness, rain, fog etc.). Therefore we will continue research on application of EC methods in the process of 2D image classification to develop the final 2D recognition system based on canny and sobel filter for SFA input images.

Summing up, presented application of EC methods to search for 2D images key-areas allows to select areas of interest. SFA is efficient and in comparison to classic methods may give better results. At the same time it allows to easily explore entire 2D input image in search for selected objects without many complicated mathematical operations like these in classic methods. It is possible to increase efficiency of SFA. If applied to filtered objects located among many other of low brightness, presented SFA may be used as dedicated recognition system for 2D input images. This is second part of the project we are working on. Early results are promising, however the idea needs more research. For example we have implemented the canny and sobel filter also in the GPU parallel version dedicated for SFA recognition. Therefore in the future research we hope to present efficient SFA recognition system for 2D input images based on GPU parallel canny and sobel filter.

\bibliographystyle{IEEEtran}

\bibliography{ieee_ssci2014_firefly_picture_keypoints}

\end{document}